\documentclass[11pt,a4paper]{article}
\usepackage[hyperref]{naaclhlt2018}
\usepackage{times}
\usepackage{latexsym}

\usepackage{url}

\usepackage{url}
\usepackage{color}
\usepackage{amsmath}
\usepackage{amssymb}
\usepackage{graphicx}
\usepackage{multirow}
\usepackage[linesnumbered,ruled,vlined]{algorithm2e}
\usepackage{algpseudocode}
\usepackage{todonotes}
\usepackage{booktabs,caption}
\usepackage[flushleft]{threeparttable}
\usepackage{textcomp}

\newcommand{\tododan}[2][]
{\todo[inline, author=Dan, color=yellow]{#2}}

\newcommand\ZZ[1]{\textrm{#1}}
\newcommand\QNfinal[1]{\textrm{#1}}
\newcommand\QN[1]{\textrm{#1}}

\newcommand{\event}[1]{\textit{\textbf{#1}}}

\newcommand{\rel}[1]{\textit{#1}}

\newcommand{\ignore}[1]{}

\newcommand{\CAEVO}{ChambersCaMcBe14}
\newcommand{\ClearTK}{Bethard13}
\newcommand{\CATENA}{MirzaTo16}
\newcounter{exctr}

\newcommand{\pad}{$\mathcal{P}$}
\newcommand{\fad}{$\mathcal{F}$}
\newcommand{\best}[1]{\textbf{#1}}
\usepackage{balance}
\aclfinalcopy % Uncomment this line for the final submission
\def\aclpaperid{103} %  Enter the acl Paper ID here

\title{Exploiting Partially Annotated Data for Temporal Relation Extraction}

\author{Qiang Ning,$^1$ Zhongzhi Yu,$^2$ Chuchu Fan,$^1$ Dan Roth$^{1,2,3}$ \\
	$^1$Department of Electrical and Computer Engineering, $^2$Department of Computer Science\\
	University of Illinois at Urbana-Champaign, Urbana, IL 61801, USA\\
	$^3$Department of Computer Science, University of Pennsylvania, Philadelphia, PA 19104, USA\\
	{\tt \small \{qning2,zyu19,cfan10\}@illinois.edu,~danroth@seas.upenn.edu}}

\date{}

\begin{document}
\maketitle
% !TEX root = sem2018.tex
\begin{abstract}
Annotating temporal relations (TempRel) between events described in natural language is known to be labor intensive, partly because the total number of TempRels is quadratic in the number of events. 
As a result, only a small number of documents are typically annotated, limiting the coverage of various lexical/semantic phenomena.
\QNfinal{In order to improve existing approaches,} one possibility is to make use of the readily available, partially annotated data ($\mathcal{P}$ \QNfinal{as in {\em partial}}) that cover more documents.
However, missing annotations in $\mathcal{P}$ are known to hurt, rather than help, existing systems.
This work is a case study in exploring various usages of $\mathcal{P}$ for TempRel extraction.
Results show that despite missing annotations, $\mathcal{P}$ is still a useful supervision signal for this task \QNfinal{within a constrained bootstrapping learning framework}.
\QNfinal{The system described in this system is publicly available.}\footnote{\url{https://cogcomp.org/page/publication_view/832}}
\end{abstract}
% !TEX root = sem2018.tex
\section{Introduction}
Understanding the temporal information in natural language text is an important NLP task \cite{VGSHKP07,VSCP10,ULADVP13,MSAAVMRUK15,BSCDPV16,BSPP17}.
A crucial component is temporal relation (TempRel; e.g., \rel{before} or \rel{after}) extraction \cite{MVWLP06,BethardMaKl07,DoLuRo12,ChambersCaMcBe14,MirzaTo16,NingFeRo17,NingFeWuRo18,NingWuPeRo18}.

The TempRels in a document or a sentence can be conveniently modeled as a graph, where the nodes are events, and the edges are labeled by TempRels.
Given all the events in an instance, TempRel annotation is the process of manually labeling all the edges -- a highly labor intensive task due to two reasons. One is that many edges require extensive reasoning over multiple sentences and labeling them is time-consuming.
Perhaps more importantly, the other reason is that \#edges is quadratic in \#nodes. If labeling an edge takes 30 seconds (already an optimistic estimation), a typical document with 50 nodes would take more than 10 hours to annotate.
Even if existing annotation schemes make a compromise by only annotating edges whose nodes are from a same sentence or adjacent sentences \cite{CassidyMcChBe14}, it still takes more than 2 hours to fully annotate a typical document.
Consequently, the only fully annotated dataset, TB-Dense \cite{CassidyMcChBe14}, contains only 36 documents, which is rather small compared with datasets for other NLP tasks.

A small number of documents may indicate that the annotated data provide a limited coverage of various lexical and semantic phenomena, since a document is usually ``homogeneous'' within itself.
In contrast to the scarcity of fully annotated datasets (denoted by $\mathcal{F}$ \QNfinal{as in {\em full}}), there are actually some partially annotated datasets as well (denoted by $\mathcal{P}$ \QNfinal{as in {\em partial}}); for example, TimeBank \cite{PHSSGSRSDF03} and AQUAINT \cite{Graff02} cover in total more than 250 documents. 
\QNfinal{Since annotators are not required to label all the edges in these datasets, it is less labor intensive to collect $\mathcal{P}$ than to collect $\mathcal{F}$.}
However, existing TempRel extraction methods only work on one type of datasets (i.e., either $\mathcal{F}$ or $\mathcal{P}$), without taking advantage of both.
No one, as far as we know, has explored ways to combine both types of datasets in learning and whether it is helpful.

This work is a case study in exploring various usages of $\mathcal{P}$ in the TempRel extraction task.
We empirically show that $\mathcal{P}$ is indeed useful within a \QN{(constrained)} bootstrapping type of learning approach.
This case study is interesting from two perspectives. 
\textbf{First}, {\em incidental supervision} \cite{Roth17}.
In practice, supervision signals may not always be perfect: they may be noisy, only partial, based on different annotation schemes, or even on different (but relevant) tasks; incidental supervision is a general paradigm that aims at making use of the abundant, naturally occurring data, as supervision signals.
As for the TempRel extraction task, the existence of many partially annotated datasets $\mathcal{P}$ is a good fit for this paradigm and the result here can be informative for future investigations involving other incidental supervision signals.
\textbf{Second}, {\em TempRel data collection}. 
The fact that $\mathcal{P}$ is shown to provide useful supervision signals poses some further questions: What is the optimal data collection scheme for TempRel extraction, fully annotated, partially annotated, or a mixture of both? For partially annotated data, what is the optimal ratio of annotated edges to unannotated edges? The proposed method in this work can be readily extended to study these questions in the future, as we further discuss in Sec.~\ref{sec:discussion}.

%, for which the label set is usually 
% !TEX root = sem2018.tex
\section{Existing Datasets and Methods}

TimeBank \cite{PHSSGSRSDF03} is a classic TempRel dataset, where the annotators were given a whole article and allowed to label TempRels between any pairs of events. 
Annotators in this setup usually focus only on salient relations but overlook some others. 
\QN{It has been reported that many event pairs in TimeBank should have been annotated with a specific TempRel but the annotators failed to look at them \cite{Chambers13,CassidyMcChBe14,NingFeRo17}.}
Consequently, we categorize TimeBank as a partially annotated dataset ($\mathcal{P}$). 
The same argument applies to other datasets that adopted this setup, such as AQUAINT \cite{Graff02}, CaTeRs \cite{CaTeRs} and RED \cite{GormanWrPa16}.
Most existing systems make use of $\mathcal{P}$, including but not limited to, \cite{MVWLP06,BDLB06,ChambersWaJu07,BethardMaKl07,VerhagenPu08,ChambersJu08,DenisMu11,DoLuRo12}; this applies also to the TempEval workshops systems, e.g., \cite{LaokulratMiTsCh13,\ClearTK,Chambers13}.

To address the missing annotation issue, \citet{CassidyMcChBe14} proposed a dense annotation scheme, TB-Dense.
\QNfinal{Edges are presented one-by-one and the annotator has to choose a label for it (note that there is a \rel{vague} label in case the TempRel is not clear or does not exist). As a result, edges in TB-Dense are considered as fully annotated in this paper.}
The first system on TB-Dense was proposed in \citet{\CAEVO}.
Two recent TempRel extraction systems \cite{\CATENA,NingFeRo17} also reported their performances on TB-Dense ($\mathcal{F}$) and on TempEval-3 ($\mathcal{P}$) separately.
However, there are no existing systems that jointly train on both.
Given that the annotation guidelines of $\mathcal{F}$ and $\mathcal{P}$ are obviously different, it may not be optimal to simply treat $\mathcal{P}$ and $\mathcal{F}$ uniformly and train on their union. This situation necessitates further investigation as we do here. 

\QNfinal{Before introducing our joint learning approach, we have a few remarks about our choice of $\mathcal{F}$ and $\mathcal{P}$ datasets.
First, we note that TB-Dense is actually not fully annotated in the {\em strict} sense because only edges within a sliding, two-sentence window are presented. That is, distant event pairs are intentionally ignored by the designers of TB-Dense. However, since such distant pairs are consistently ruled out in the training and inference phase in this paper, it does not change the nature of the problem being investigated here. At this point, TB-Dense is the only fully annotated dataset that can be adopted in this study, despite the aforementioned limitation.}

\QNfinal{Second, the partial annotations in datasets like TimeBank were not selected  uniformly at random from all possible edges. As described earlier, only salient and non-vague TempRels (which may often be those easy ones) are labeled in these datasets. Using TimeBank as \pad{} might potentially create some bias and we will need to keep this in mind when analyzing the results in Sec.~\ref{sec:result}.
Recent advances in TempRel data annotation \cite{NingWuRo18} can be used in the future to collect both $\mathcal{F}$ and $\mathcal{P}$ more easily.}

\ignore{
\tododan{the next paragraph seems out of flow; needed? Maybe you can link by saying that one way to better exploit the additional data is to enforce constraints during joint learning and then talk about the constraints.}
The transitivity rules,\footnote{For instance, if \event{e1} is \rel{before} \event{e2} and \event{e2} is \rel{before} \event{e3}, then \event{e1} must be \rel{before} \event{e3}.} which a temporal graph should follow, are usually applied as constraints in the {\em global} inference step (i.e., the step that generates the output graphs). The global inference can be solved greedily \cite{\CAEVO,\CATENA} or exactly \cite{BDLB06,ChambersJu08,DenisMu11,DoLuRo12,NingFeRo17}.
}

% !TEX root = sem2018.tex
\section{Joint Learning on $\mathcal{F}$ and $\mathcal{P}$}
\label{sec:proposed}
In this work, we study two learning paradigms that make use of both $\mathcal{F}$ and $\mathcal{P}$.
In the first, we simply treat those edges that are annotated in \pad{} as edges in \fad{} so that the learning process can be performed on top of the union of \fad{} and \pad{}.
This is the most straightforward approach to using $\mathcal{F}$ and $\mathcal{P}$ jointly and it is interesting to see if it already helps.

In the second, we use bootstrapping: we use $\mathcal{F}$ as a starting point and learn a TempRel extraction system on it (denoted by $S_\mathcal{F}$), and then fill those missing annotations in \pad{} based on $S_\mathcal{F}$ (thus obtain ``fully'' annotated $\tilde{\mathcal{P}}$); finally, we treat $\tilde{\mathcal{P}}$ as \fad{} and learn from both. Algorithm~\ref{algo:bootstrap} is a meta-algorithm of the above.
\begin{algorithm}
	\DontPrintSemicolon % Some LaTeX compilers require you to use \dontprintsemicolon instead
	\KwIn{\fad{}, \pad{}, Learn, Inference}
	$S_\mathcal{F}=\textrm{Learn}(\mathcal{F})$ \label{ln:learn1}\;
	Initialize $S_{\mathcal{F}+\mathcal{P}}=S_\mathcal{F}$\;
	\While{convergence criteria not satisfied}{
		$\tilde{\mathcal{P}}=\emptyset$\;
		\ForEach{$p\in\mathcal{P}$}{
			$\hat{\mathbf{y}}$ = Inference($p;S_{\mathcal{F}+\mathcal{P}}$)\label{ln:inference}\;
			$\tilde{\mathcal{P}}=\tilde{\mathcal{P}}\cup \{(\mathbf{x},\hat{\mathbf{y}})\}$\;
		}
		$S_{\mathcal{F}+\mathcal{P}} =\textrm{Learn}(\mathcal{F}+\tilde{\mathcal{P}})$\label{ln:learn2}\;
	}
	\Return{$S_{\mathcal{F}+\mathcal{P}}$}\;
	\caption{Joint learning from \fad{} and \pad{} by bootstrapping}
	\label{algo:bootstrap}
\end{algorithm}

\begin{figure}[htbp!]
	\centering
	\includegraphics[width=0.3\textwidth]{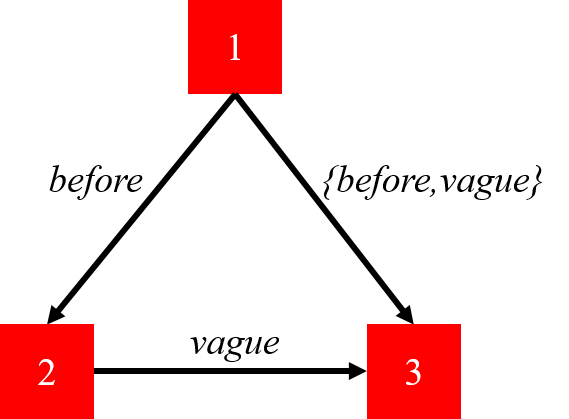}
	\caption{\small Nodes 1-3 are three time points and let $(i,j)$ be the edge from node $i$ to node $j$, where $(i,j)\in$\{before, after, equal, vague\}. Assume the current annotation is $(1,2)=before$ and $(2,3)=vague$ and $(1,3)$ is missing. However, $(1,3)$ cannot be \rel{after} because it leads to $(2,3)=after$, conflicting with their current annotation; similarly, $(1,3)$ cannot be \rel{equal}, either.}
	\label{fig:partial effect}
\end{figure}

In Algorithm~\ref{algo:bootstrap}, we consistently use the sparse averaged perceptron algorithm as the ``Learn'' function.
As for ``Inference'' (Line~\ref{ln:inference}), we further investigate two different ways: (i) Look at every unannotated edge in $p\in\mathcal{P}$ and use $S_{\mathcal{F}+\mathcal{P}}$ to label it; this {\em local} method ignores the existing annotated edges in $\mathcal{P}$ and \QN{is thus the {\em standard} bootstrapping. (ii) Perform global inference on $\mathcal{P}$ with annotated edges being constraints, which is a {\em constrained} bootstrapping,} motivated by the fact that temporal graphs are structured and annotated edges have influence on the missing edges:
In Fig.~\ref{fig:partial effect}, \QNfinal{the current annotation for $(1,2)$ and $(2,3)$ is \rel{before} and \rel{vague}. We assume that the annotation $(2,3)$=\rel{vague} indicates that the relation cannot be determined even if the entire graph is considered. Then with $(1,2)$=\rel{before} and $(2,3)$=\rel{vague}, we can see that $(1,3)$ cannot be uniquely determined, but it is restricted to be selected from $\{before, vague\}$ rather than the entire label set.}
We believe that global inference makes better use of the information provided by \pad{}; \ZZ{in fact}, as we show in Sec.~\ref{sec:result}, it does perform better than local inference.

A standard way to perform global inference is to formulate it as an Integer Linear Programming (ILP) problem\cite{RothYi04} and enforce transitivity rules as constraints. 
%Assume the existence of a classifier that produces soft-max scores for $(i,j)=r$. 
Let $\mathcal{R}$ be the TempRel label set\footnote{In this work, we adopt \rel{before}, \rel{after}, \rel{includes}, \rel{be\_included}, \rel{simultaneously}, and \rel{vague}.}, $\mathcal{I}_r(ij)\in\{0,1\}$ be the indicator function of $(i,j)=r$, and $f_r(ij)\in[0,1]$ be the corresponding soft-max score obtained via $\mathcal{S}_{\mathcal{F+P}}$. Then the ILP objective is formulated as
%in Eq.~\eqref{eq:basic ilp}.
\begin{eqnarray}
	\label{eq:basic ilp}
	&\hat{\mathcal{I}} = \textrm{arg}\underset{\mathcal{I}}{\textrm{max}}\sum_{i<j}\sum_{r\in\mathcal{R}} f_r(ij) \mathcal{I}_r(ij)\\
	&\textrm{s.t.}\quad\underset{\textrm{(uniqueness)}}{\Sigma_{r}{\mathcal{I}_r(ij)} = 1},\nonumber\\
	&\underset{\textrm{(transitivity)}}{\mathcal{I}_{r_1}(ij)+\mathcal{I}_{r_2}(jk)-\Sigma_{m=1}^N \mathcal{I}_{r_{3}^m}(ik)\le 1,}\nonumber
\end{eqnarray}
where $\{r_3^m\}$ is selected based on the general transitivity proposed in \cite{NingFeRo17}.
With Eq.~\eqref{eq:basic ilp}, different implementations of Line~\ref{ln:inference} in Algorithm~\ref{algo:bootstrap} can be described concisely as follows: (i) Local inference is performed by ignoring ``transitivity constraints''. (ii) Global inference can be performed by adding annotated edges in $\mathcal{P}$ as additional constraints.
Note that Algorithm~\ref{algo:bootstrap} is only for the learning step of TempRel extraction; as for the inference step of this task, we \QN{consistently adopt the standard method by solving Eq.~\eqref{eq:basic ilp}, as was done by \cite{BDLB06,ChambersJu08,DenisMu11,DoLuRo12,NingFeRo17}.}

% constraints vs. scores
% 1-model vs. 2-model
% direct put P into F vs. codl (no use labels in P) vs.  codl (use labels in P)
% add TBAQ to P
% compare to EMNLP
% F1 vs. F_awareness
% downsampling P
% testset from P
% only consider within-sentence pairs
% !TEX root = sem2018.tex
\section{Experiments}
\label{sec:result}
In this work, we consistently used TB-Dense as the fully annotated dataset (\fad{}) and TBAQ as the partially annotated dataset (\pad{}).
The corpus statistics of these two datasets are provided in Table~\ref{tab:stat}.
Note that TBAQ is the union of TimeBank and AQUAINT and it originally contained 256 documents, but 36 out of them completely overlapped with TB-Dense, so we have excluded these when constructing \pad{}.
In addition, the number of edges shown in Table~\ref{tab:stat} only counts the event-event relations (i.e., do not consider the event-time relations therein), which is the focus of this work.

\begin{table}[htbp!]
	\small
	\begin{tabular}{c|c|c|c|c}
		\hline
		Data&\#Doc&\#Edges&Ratio&Type\\
		\hline
		TB-Dense&36&6.5K&100\%&\fad{}\\
		TBAQ&220&2.7K&12\%&\pad{}\\
		\hline
	\end{tabular}
	\centering
	\caption{\small Corpus statistics of the fully and partially annotated dataset used in this work. TBAQ: The union of TimeBank and AQUAINT, which is the training set provided by the TempEval3 workshop. \QNfinal{\#Edges: The number of annotated edges.} \QN{Ratio: The proportion of annotated edges.}}
	\label{tab:stat}
\end{table}

We also adopted the original split of TB-Dense (22 documents for training, 5 documents for development, and 9 documents for test). Learning parameters were tuned to maximize their corresponding F-metric on the development set. Using the selected parameters, systems were retrained with development set incorporated and evaluated against the test split of TB-Dense (about 1.4K relations: 0.6K \rel{vague}, 0.4K \rel{before}, 0.3K \rel{after}, and 0.1K for the rest). Results are shown in Table~\ref{tab:result}, where all systems were compared in terms of their performances on ``same sentence" edges (both nodes are from the same sentence), ``nearby sentence'' edges, all edges, \QN{and the temporal awareness metric} used by the TempEval3 workshop.

\begin{table*}[htbp!]
	\small
	\begin{tabular}{ c|l|c|ccc|ccc|ccc|ccc } 
		\hline
		\multirow{2}{*}{No.}&\multicolumn{2}{c|}{Training}&\multicolumn{3}{c|}{\textit{Same Sentence}}&\multicolumn{3}{c|}{\textit{Nearby Sentence}}&\multicolumn{3}{c|}{\best{Overall}}&\multicolumn{3}{c}{\best{Awareness}}\\
		\cline{2-15}&Data&Bootstrap&P&R&F&P&R&F&P&R&F&P&R&F\\
		\hline
		1&	\fad{}	&-&{47.1}&{49.7}&{48.4}&{40.2}&37.9&39.0&\best{42.1}&41.0&\best{41.5}&\best{40.0}&40.7&\best{40.3}\\
		2&	$\mathcal{P}^{Full}$&-&37.0&33.1&35.0&34.4&19.6&24.9&37.7&23.6&29.0&36.9&24.0&29.1\\
		3&	\pad{}	&-&34.1&52.5&41.3&26.1&{48.1}&33.8&30.2&\best{52.1}&38.2&28.6&\best{49.9}&36.4\\
		4&	\fad{}+$\mathcal{P}^{Full}$&-&38.5&32.2&35.1&40.1&38.1&{39.1}&40.8&35.3&37.8&37.1&36.2&36.6\\
		5&	\fad{}+\pad{}&-&43.7&43.9&43.8&39.1&38.3&38.7&41.8&40.7&41.2&38.6&41.4&40.0\\
		\hline\hline
		6&	\fad{}+$\mathcal{P}^{Empty}$&Local&41.7&50.3&45.6&39.5&48.1&43.4&41.8&50.4&45.7&40.9&47.5&43.9\\
		7&	\fad{}+$\mathcal{P}^{Empty}$&Global&{44.7}&{55.5}&{49.5}&{40.1}&{48.7}&{44}&\best{42.0}&\best{51.4}&\best{46.2}&\best{41.1}&\best{48.3}&\best{44.4}\\
		\hline\hline
		8&	\fad{}+\pad{}&Local&43.6&50&46.6&43&46.9&44.8&43.7&47.8&45.6&42&45.6&43.7\\
		9&	\fad{}+\pad{}&Global&{44.9}&{56.1}&{49.9}&{43.4}&{52.3}&{47.5}&\best{44.7}&\best{54.1}&\best{49.0}&\best{44.1}&\best{50.8}&\best{47.2}\\
		\hline
		\ignore{10&\citet{NingFeRo17}&Global&46.6&44.5&45.5&41.1&47.7&44.2&43.0&46.4&44.7&42.6&44.0&43.3\\
		\hline}
	\end{tabular}
	\centering
	\caption{\small Performance of various usages of the partially annotated data in training. \fad{}: Fully annotated data. \pad{}: Partially annotated data. $\mathcal{P}^{Full}$: \pad{} with missing annotations filled by \rel{vague}. $\mathcal{P}^{Empty}$: \pad{} with all annotations removed. Bootstrap: referring to specific implementations of Line~\ref{ln:inference} in Algorithm~\ref{algo:bootstrap}, i.e., local or global. Same/nearby sentence: edges whose nodes appear in the same/nearby sentences in text. Overall: all edges. Awareness: the temporal awareness metric used in the TempEval3 workshop, measuring how useful the predicted graphs are \cite{ULADVP13}. \QN{System~7 can also be considered as a reproduction of \citet{NingFeRo17} (see the discussion in Sec.~\ref{sec:discussion} for details).}}
	\label{tab:result}
\end{table*}

The first part of Table~\ref{tab:result} (Systems~1-5) refers to the baseline method proposed at the beginning of Sec.~\ref{sec:proposed}, i.e., simply treating \pad{} as \fad{} and training on their union. $\mathcal{P}^{Full}$ is a variant of \pad{} by filling its missing edges by \rel{vague}. Since it labels too many \rel{vague} TempRels, System~2 suffered from a low recall.
In contrast, \pad{} does not contain any \rel{vague} training examples, so System~3 would only predict specific TempRels, leading to a low precision.
Given the obvious difference in \fad{} and $\mathcal{P}^{Full}$, System~4 expectedly performed worse than System~1.
However, when we see that System~5 was still worse than System~1, it is surprising because the annotated edges in \pad{} are correct and should have helped.
This unexpected observation suggests that simply adding the annotated edges from \pad{} into \fad{} is not a proper approach to learn from both.

The second part (Systems~6-7) serves as an ablation study showing the effect of bootstrapping only.  $\mathcal{P}^{Empty}$ is another variant of \pad{} we get by removing all the annotated edges (that is, only nodes are kept). Thus, they did not get any information from the annotated edges in \pad{} and any improvement came from bootstrapping alone. \QN{Specifically, System~6 is the standard bootstrapping and System~7 is the constrained bootstrapping.}

Built on top of Systems~6-7, Systems~8-9 further took advantage of the annotations of \pad{}, which resulted in additional improvements.
Compared to System 1 (trained on \fad{} only) and System 5 (simply adding \pad{} into \fad{}), the proposed System~9 achieved much better performance, which is also statistically significant with p$<$0.005 (McNemar's test). 
\QNfinal{While System 7 can be regarded as a reproduction of \citet{NingFeRo17}, the original paper of \citet{NingFeRo17} achieved an overall score of P=43.0, R=46.4, F=44.7 and an awareness score of P=42.6, R=44.0, and F=43.3,} and the proposed System~9 is also better than \citet{NingFeRo17} on all metrics.\footnote{We obtained the original event-event TempRel predictions of \citet{NingFeRo17} from \url{https://cogcomp.org/page/publication_view/822}.} 
% !TEX root = sem2018.tex
\section{Discussion}
\label{sec:discussion}

\ignore{
We notice that the proposed Systems 8-9 mainly improves over System 1 in terms of recall, which may be because \pad{} is not a random partial annotation, but a partial annotation with only non-vague relations; such a bias in \pad{} is already reflected by the high recall of System~3 and may be the reason for the high recall of the proposed systems as well.
}

While incorporating transitivity constraints in inference is widely used, \citet{NingFeRo17} proposed to incorporate these constraints in the learning phase as well.
One of the algorithms proposed in \citet{NingFeRo17} is based on~\citet{ChangRaRo12}'s constraint-driven learning (CoDL), which is the same as our intermediate System 7 in Table~\ref{tab:result}; the fact that System 7 is better than System 1 can thus be considered as a reproduction of \citet{NingFeRo17}.
Despite the technical similarity, this work is motivated differently and is set to achieve a different goal: \citet{NingFeRo17} tried to enforce the transitivity structure, while the current work attempts to use imperfect signals (e.g., partially annotated) taken from additional data, and learn in the incidental supervision framework.

The \pad{} used in this work is TBAQ, where only 12\% of the edges are annotated. 
In practice, every annotation comes at a cost, either time or the expenses paid to annotators, and as more edges are annotated, the marginal ``benefit'' of one edge is going down (an extreme case is that an edge is of no value if it can be inferred from existing edges).
Therefore, a more general question is to find out the optimal ratio of graph annotations.

Moreover, partial annotation is only one type of annotation imperfection. If the annotation is noisy,  
we can alter the hard constraints derived from \pad{} and use soft regularization terms; if the annotation is for a different but relevant task, we can formulate corresponding constraints to connect that different task to the task at hand.
Being able to learn from these ``indirect'' signals is appealing because indirect signals are usually order of magnitudes larger than datasets dedicated to a single task.

% !TEX root = sem2018.tex
\section{Conclusion}

Temporal relation (TempRel) extraction is important but TempRel annotation is labor intensive.
While fully annotated datasets (\fad{}) are relatively small, there exist more datasets with partial annotations (\pad{}).
This work provides the first investigation of learning from both types of datasets, and this preliminary study already shows promise. 
\QN{Two bootstrapping algorithms (standard and constrained) are analyzed} and the benefit of \pad{}, although with missing annotations, is shown on a benchmark dataset.
This work may be a good starting point for further investigations of incidental supervision and data collection schemes of the TempRel extraction task.

\section*{Acknowledgements}

We thank all the reviewers for providing insightful comments and critiques. This research is supported in part by a grant from the Allen Institute for Artificial Intelligence (allenai.org); the IBM-ILLINOIS Center for Cognitive Computing Systems Research (C3SR) - a research collaboration as part of the IBM AI Horizons Network; by DARPA under agreement number FA8750-13-2-0008; and by the Army Research Laboratory (ARL) under agreement W911NF-09-2-0053.

The U.S. Government is authorized to reproduce and distribute reprints for Governmental purposes notwithstanding any copyright notation thereon. 
The views and conclusions contained herein are those of the authors and should not be interpreted as necessarily representing the official policies or endorsements, either expressed or implied, of DARPA or the U.S. Government.
Any opinions, findings, conclusions or recommendations are those of the authors and do not necessarily reflect the view of the ARL.

\bibliography{sem2018,kbcom18,cited-long,ccg-long}
\bibliographystyle{acl_natbib.bst}

\end{document}